\documentclass{article}
\usepackage{ijcai25}

\usepackage{times}
\usepackage{soul}
\usepackage{url}
\usepackage[hidelinks]{hyperref}
\usepackage[utf8]{inputenc}
\usepackage[small]{caption}
\usepackage{graphicx}
\usepackage{amsmath}
\usepackage{amsthm}
\usepackage{booktabs}
\usepackage{algorithm}
\usepackage{algorithmic}
\usepackage[switch]{lineno}
\usepackage{array} 
\usepackage{multirow}
\usepackage{colortbl}
\usepackage{xcolor}
\usepackage{amssymb}

\newtheorem{theorem}{Theorem}

\title{Multi-Omics Analysis for Cancer Subtype Inference via Unrolling Graph Smoothness Priors}

\author{
Jielong Lu
\and
Zhihao Wu
\and
Jiajun Yu
\and
Jiajun Bu
\and
Haishuai Wang\thanks{Corresponding author: Haishuai Wang}
\affiliations
Zhejiang Key Laboratory of Accessible Perception and Intelligent Systems, \\College of Computer Science and Technology,	Zhejiang University\\
\emails
jielonglu2022@163.com,
zhihaowu1999@gmail.com,
jiajunyu1999@gmail.com,\\
bjj@zju.edu.cn,
haishuai.wang@zju.edu.cn
}

\begin{document}

\maketitle

\begin{abstract}
Integrating multi-omics datasets through data-driven analysis offers a comprehensive understanding of the complex biological processes underlying various diseases, particularly cancer.
Graph Neural Networks (GNNs) have recently demonstrated remarkable ability to exploit relational structures in biological data, enabling advances in multi-omics integration for cancer subtype classification. 
Existing approaches often neglect the intricate coupling between heterogeneous omics, limiting their capacity to resolve subtle cancer subtype heterogeneity critical for precision oncology.
To address these limitations, we propose a framework named Graph Transformer for Multi-omics Cancer Subtype Classification (GTMancer).
This framework builds upon the GNN optimization problem and extends its application to complex multi-omics data.
Specifically, our method leverages contrastive learning to embed multi-omics data into a unified semantic space.
We unroll the multiplex graph optimization problem in that unified space and introduce dual sets of attention coefficients to capture structural graph priors both within and among multi-omics data.
This approach enables global omics information to guide the refining of the representations of individual omics.
Empirical experiments on seven real-world cancer datasets demonstrate that GTMancer outperforms existing state-of-the-art algorithms.
\end{abstract}

\section{Introduction}
With the rapid development of biomedical technology, more and more omics data can be obtained, such as genomics, transcriptomics, proteomics, etc.
Data-driven multi-omics analyses in biomedicine enable researchers to comprehensively understand the key biological processes underlying diseases \cite{stephenson2021single,theodoris2023transfer,wang2024marsgt,lee2024unraveling}.
Given the inherent complexity and multilayered regulatory mechanisms of biological systems, single-omics studies often struggle to elucidate the relationships between molecular alterations and phenotypic traits comprehensively \cite{karczewski2018integrative}.
Many diseases, including cancer, result from multi-stage processes that integrate multi-scale information spanning from the genome to the proteome \cite{graph_multi_omics}.
Consequently, multi-omics analysis enables a more comprehensive exploration of interactions and synergistic effects.  
While current multi-omics integration approaches have yielded certain advancements, a notable drawback is their inability to comprehensively capture the intricate relationships within omics data.
Graph-based multi-omics learning approaches construct graphs for individual modalities by treating the samples within each modality as graph nodes and representing their relationships through edges \cite{lewis2021integration,tsai2023histopathology,zheng2024global}.
This approach facilitates the development of a well-defined structured representation within each modality, enabling graph-based model to more effectively capture feature interactions and underlying patterns specific to the modality \cite{schulte2021integration,fang2021deepan,Li2022MOGCN,wu2024mosgat}.

Graph Neural Networks (GNNs)  effectively leverage graph-based prior knowledge, such as protein interaction networks, attracting significant interest from researchers~\cite{GCN,li2021structure,mastropietro2023learning,wu2024graph,fang2025insgnn,fang2025large}.
Cancer multi-omics data lack explicitly provided structural priors, requiring considerable effort and domain expertise to create graphs that capture the underlying biological relationships.
A common approach in GNN-based cancer multi-omics research involves representing each omics dataset as an individual graph, which is subsequently fused into a unified homogeneous graph.
This unified graph is then used as input for GNNs to perform tasks such as subtype classification or survival prediction \cite{li2022graph,wu2024mosgat,gao2025precision}.
Alternatively, the independent graphs can be processed using multi-channel GNNs, with the final fused representation obtained at the semantic layer to be utilized for subsequent tasks.
Specifically, these methods typically process each omics modality independently during message passing, resulting in fragmented representations that fail to capture the rich interdependencies among omics. 
This siloed treatment limits comprehensive cross-omics interactions and hinders the real-time sharing of global biological information. Consequently, updates to multi-omics embeddings may deviate from directions that best reflect the underlying cancer heterogeneity, restricting model expressiveness and downstream classification accuracy for cancer subtypes.

To address these challenges, we revisit GNNs through an optimization-inspired lens, recognizing that many popular architectures can be interpreted as iterative solutions to smoothness-based objectives via gradient descent. Extending this perspective to multi-omics data, we propose a framework that aligns heterogeneous omics profiles into a unified semantic space through contrastive learning.
This alignment facilitates meaningful information exchange across modalities, which is critical for capturing the complex omics interplay driving cancer subtype diversity.
After alignment, we formulate a graph optimization problem for multi-omics analysis, aiming to ensure that representations of the same omics across different cancer patients gradually converge, while representations of different omics within the same cancer patient also become similar. 
To prevent the updating process from deviating prematurely, we incorporate a regularization term.
In this optimization problem, we design two sets of attention coefficients to represent the graph structural priors within and among omics data, respectively. 
By refining the optimization framework, we derive a preliminary update formula. A key insight from our analysis is the critical role of step size in gradient-based optimization; improper tuning can cause slow convergence or even divergence, especially given the heterogeneity of multi-omics cancer data. To overcome this, we propose leveraging Newton’s method to solve the optimization problem, obviating the need for manual step size selection and guaranteeing stable, theoretical convergence.
Our contributions are summarised as follows:
\begin{itemize}
    \item We propose GTMancer, an optimization-inspired framework that enables effective global integration across multi-omics modalities for cancer subtype classification.
    \item We provide theoretical evidence demonstrating that an artificially defined step size is unnecessary, as GTMancer achieves stable convergence through iterations.
    \item The proposed method demonstrates strong performance in comparison to other state-of-the-art algorithms across seven real-world cancer datasets.
\end{itemize}

\section{Related Work}
\subsection{Graph-based Multi-omics Analysis}
Graph-based multi-omics integrative analysis promotes new research by combining patient information and biomedical knowledge. Several studies have demonstrated the potential of graph-based approaches in multi-omics integration, providing inspiration and insights to scientists and clinicians for addressing their carefully designed research questions.
\citeauthor{pai2019netdx}  proposed to utilize the patient similarity graph, which improves survival prediction across four tumor types and enhances result interpretability by visualizing decision boundaries in the patient similarity space \cite{pai2019netdx}.
\citeauthor{fang2021deepan} demonstrated the effectiveness of learning patient similarity features using a marginal graph autoencoder, followed by graph clustering, to stratify non-small cell lung cancer patients into subgroups with different immunotherapy outcomes \cite{fang2021deepan}.
Given the strong performance of graph neural networks across various domains, many researchers have applied them to multi-omics analysis.
\citeauthor{schulte2021integration} proposed an interpretable machine learning method based on graph convolutional networks that predicted cancer genes by integrating multi-omics pan-cancer data (e.g., mutations, copy number variations, DNA methylation, and gene expression) with protein-protein interaction (PPI) networks \cite{schulte2021integration}.
\citeauthor{Li2022MOGCN} applied a multimodal autoencoder to extract features and used a similarity network fusion model to build a patient similarity network, which, through a graph convolutional network, integrated these heterogeneous features to train a subtype classification model \cite{Li2022MOGCN}.
\citeauthor{wang2024marsgt} conducted a multi-omics analysis of rare cell population inference using a single-cell graph transformer, which identifies rare populations through a probability-based heterogeneous graph model on multi-omics data, providing biological insights for both synthetic and real datasets \cite{wang2024marsgt}.
Existing approaches struggle to integrate omics data into a cohesive narrative and fail to account for the complex interrelationships between molecular entities while considering connectivity patterns across multiple histological collections.

\subsection{Graph Neural Networks}
Graph neural networks have garnered significant attention due to their ability to simultaneously model both topology and node features. 
With the advancement of biotechnology, graph neural networks are increasingly being applied in this field.
\citeauthor{wang2021scgnn} employed graph neural networks to model and aggregate cell-cell relationships, and used a left-truncated Gaussian mixture model to capture heterogeneous gene expression patterns \cite{wang2021scgnn}.
\citeauthor{wen2022graph} proposed the general graph neural network framework scMoGNN to model the relationships between modalities and integrate large unimodal datasets into downstream analyses \cite{wen2022graph}.
\citeauthor{ma2023single} modeled scMulti-omics in a heterogeneous graph and used a multi-headed graph transformer to learn the relationships between cells and genes in local and global contexts in a robust manner \cite{ma2023single}.
\citeauthor{liu2024muse} developed an informative graph structure for model training and gene representation generation, combining regularization, weighted similarity learning, and contrastive learning to capture gene-gene relationships across data \cite{liu2024muse}.
 \citeauthor{fan2024scgraphformer} moved beyond relying on predefined graphs by learning a comprehensive network of cell-cell relationships directly from scRNA-seq data, constructing a dense graph structure that captured the full range of cellular interactions \cite{fan2024scgraphformer}.
These methods are often restricted to interactions within a single omics, overlooking the  impact of interactions between different omics of information.

\begin{figure*}[!htbp] \label{framework}
	\centering
	\includegraphics[width=0.97\textwidth]{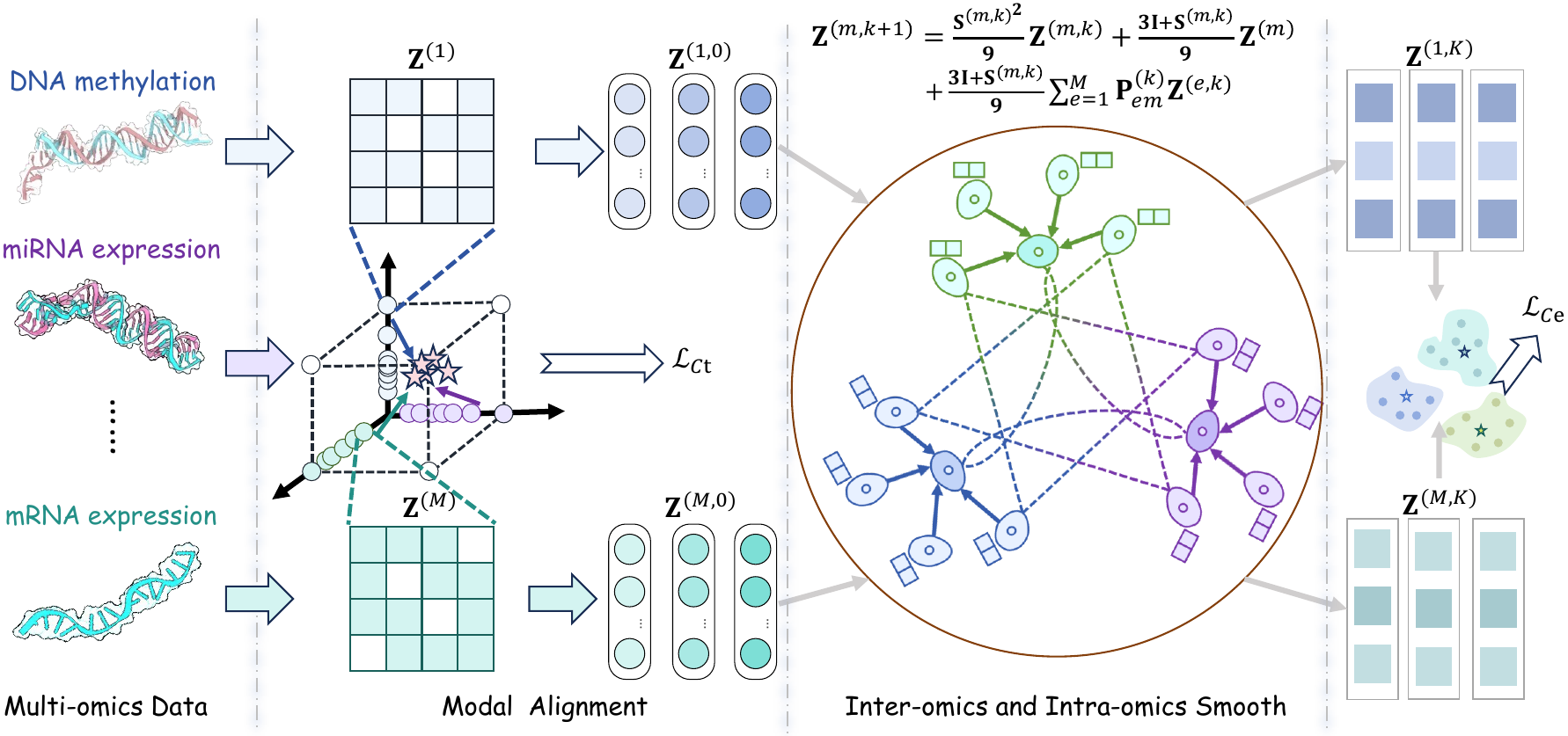}\\
	\caption{Proposed GTMancer Framework. The framework dynamically models interactions between different histological modalities. First, semantic alignment across multiple omics modalities is achieved using contrastive learning. Following alignment, dynamic interactions between modalities are performed to capture global information across modalities. Finally, the latent representation is refined through $K$ iterations of these interactions.}
\end{figure*}

\section{Method}
In this section, we first review GNNs from an optimization perspective and subsequently extend them to the multi-omics domain. 
By applying optimization techniques, we derive the forward propagation formula, which updates the representation of each omics dataset.
This process integrates attention mechanisms both within and between omics modalities, thereby enhancing the representations for downstream tasks.
\subsection{Preliminary}
Denote given multi-omics data as $\{\mathbf{V}^{(m)} \in \mathbb{R}^{N \times D^m} \}^{M}_{m=1}$, where $\{\mathbf{V}^{(1)},\cdots, \mathbf{V}^{(M)}\}$ denotes distinct modalities, $N$ is the number of samples, and $D^m$ is the dimension of the $m$-th omics.
Let $\mathbf{Y} \in \mathbb{R}^{N \times c}$ be the label matrix and $c$ denotes the number of classes.
$||\cdot||_2$ denotes the Euclidean norm of vector and $||\cdot||_F$ represents the Frobenius norm of matrix.
\subsection{Proposed Method}
 We first revisit the forward propagation formulation of the GNN, it is equivalent to the stepwise optimization of an objective function that captures the interactions between node features. This objective function is given by:
\begin{equation} 
	\mathcal{H}(\mathbf{F})= \frac{1}{2}\sum_{i,j}\mathbf{A}_{ij}\|\mathbf{F}_i-\mathbf{F}_j\|_2^2,
\end{equation}
where $\mathbf{F}$ is the feature matrix, $\mathbf{A}$ is the adjacency matrix of the graph.  
By minimizing this objective function, we encourage connected nodes to have similar feature representations, thereby promoting smoothness across the graph.
 To optimize $\mathcal{H}(\mathbf{F})$, we compute its gradient with respect to the node features $\mathbf{F}$. The gradient is derived as follows:
\begin{equation} 
	\triangledown\mathcal{H}(\mathbf{F})=(\mathbf{I}-\mathbf{A}) \mathbf{F},
\end{equation}
Utilizing the gradient, we update the node features iteratively through the following update rule:
  \begin{equation} 
	\mathbf{F}^{(k+1)} =\mathbf{F}^{(k)}- \alpha\triangledown\mathcal{H}(\mathbf{F}^{(k)})=(1-\alpha)\mathbf{F}^{(k)} + \alpha \mathbf{A}\mathbf{F}^{(k)},
\end{equation}
When $\alpha=0$, $\mathbf{F}^{(k+1)} = \mathbf{F}^{(k)}$, the model only performs a linear transformation of node features, which is equivalent to an MLP.
When $0<\alpha<1$, the model is equivalent to an extended version of APPNP with residual connections. By adjusting $\alpha$, a balance can be found between feature transformation and neighborhood aggregation.
When $\alpha=1$, $\mathbf{F}^{(k+1)} = {\mathbf{A}}\mathbf{F}^{(k)}$, the model degenerates to a standard GCN, which completes feature aggregation by directly averaging neighborhood information.

Since data from different omics reside in distinct semantic spaces, effectively exploring their interactions requires robust encoding methods for seamless multi-modal collaboration. 
To achieve this, we focus on developing such encoding strategies. 
Building on previous research, we adopt a contrastive approach to align data from various biological modalities, such as DNA methylation, mRNA expression, and protein levels, aiming to reconcile their differences and emphasize shared biological patterns as follows:

\begin{equation} \label{init}
	\mathbf{Z}^{(m)} = \mathbf{V}^{(m)}\mathbf{W}^{(m)}+\mathbf{b} ^{(m)},
  \end{equation}
where $\mathbf{W}^{(m)} \in \mathbb{R}^{D^m \times d}$ and $\mathbf{b} ^{(m)} \in \mathbb{R}^{d}$ are the trainable weight matrices and bias.
To align the encoded representations across multiple omics modalities, we employ a normalized similarity measure:
\begin{equation}
   \mathbf{\Gamma^{(m)}} = \frac{\mathbf{Z}^{(m)}}{\|\mathbf{Z}^{(m)}\|_F}\left(\frac{\overline{\mathbf{Z}}}{\|\overline{\mathbf{Z}}\|_F}\right)^\top\cdot\exp(\tau)
\end{equation}
where $\overline{\mathbf{Z}}= \frac{1}{M}\sum_{m=1}^M\mathbf{Z}^{(m)}$ represents the aggregated representation across all modalities. The temperature parameter $\tau$ controls the concentration level of the similarity distribution, ensuring robust gradient signals during optimization.
The contrastive loss is formulated to maximize the agreement between the aligned representations of different omics modalities, encouraging the model to learn consistent and biologically coherent embeddings across these heterogeneous data types.
\begin{equation}
    \mathcal{L}_{\mathrm{Ct}}=-\frac{1}{M}\sum_{m=1}^M\frac{1}{N}\sum_{i,j=1}^N\mathbf{T}_{ij}(\log\mathbf{\Gamma}^{(m)}_{ij}+\log(\mathbf{\Gamma}^{(m)\top})_{ij}),
\end{equation}
where $\mathbf{T} = \mathbb{R}^{N\times N}$ is the label matrix. The two terms in the summation enforce symmetric consistency between modalities.
Following the alignment step, the data from different omics modalities are projected into a shared semantic space.
Building upon this representation, we extend the classic graph-smoothness objective so that it simultaneously models sample-level and modality-level relations important for distinguishing subtle cancer subtypes:
\begin{equation}
\begin{aligned}
    \mathcal{H}&(\{\mathbf{Z}^{(m)}\}^{M}_{m=1}) = 
\underbrace{\frac{1}{2}\sum_{m=1}^M\sum_{i,j =1}^N \mathbf{S}^{(m)}_{ij}\|\mathbf{Z}_i^{(m)}- \mathbf{Z}_j^{(m)}\|_2^2 }_{\text{Intra-omics smoothness}}+ \\
    & \underbrace{\frac{1}{4}\sum_{e,n = 1}^M\mathbf{P}_{en}\|\mathbf{Z}^{(e)}- \mathbf{Z}^{(n)}\|_F^2}_{\text{Inter-omics smoothness}} + \underbrace{\frac{1}{2}\sum_{m=1}^M\|\mathbf{Z}^{(m)} - \mathbf{Z}^{(m)}_{init}\|_F^2}_{ \text{Regularization}},\\
    &\text{s.t.}~~ \mathbf{S}^{(m)} = \mathbf{S}^{(m)\top} , ~\mathbf{P}\mathbf{1} = \mathbf{1}, ~\mathbf{P} = {\mathbf{P}}^\top.
\end{aligned}
\end{equation}
where $\mathbf{S}^{(m)}_{ij}$ denotes the similarity between the $i$-th sample and the $j$-th sample in the $m$-th omics, $\mathbf{P}_{en}$ denotes the similarity between the $e$-th omics and the $n$-th omics, and $\mathbf{Z}^{(m)}_{init}$ is $\mathbf{Z}^{(m)}$ in Equation \eqref{init}.

The first term enforces smoothness within each omics, ensuring that patients with similar profiles are embedded closely in the latent space, which helps retain modality-specific biological signals.
The second term encourages coherence between different omics, such as aligning the representations of DNA methylation and protein expression for the same patient, which facilitates integrated modeling of complementary biological information crucial for understanding cancer heterogeneity.
The third term acts as a regularizer, preventing drastic changes from the initial modality-specific embeddings, thereby maintaining biologically meaningful baseline signals during the integration process.
To optimize $\mathcal{H}(\{\mathbf{Z}^{(m)}\}^{M}_{m=1})$, we compute its gradient with respect to the node features $\mathbf{Z}^{(m)}$. The gradient is derived as follows:
\begin{equation}
\begin{aligned}
   \triangledown \mathcal{H}(\mathbf{Z}^{(m)}) = (3\mathbf{I}-\mathbf{S}^{(m)})\mathbf{Z}^{(m)} - \sum_{e=1}^M  \mathbf{P}_{em}\mathbf{Z}^{(e)} -\mathbf{Z}^{(m)}_{init},   \
\end{aligned}
\end{equation}
where $\mathbf{I}$ represents the identity matrix.
Then,  each iterative step with step size $\alpha$ can be formulated as the first-order gradient descent:
 \begin{equation} \label{first-order}
 \begin{aligned}
    &\mathbf{Z}^{(m,k+1)} =\mathbf{Z}^{(m,k)}- \alpha\triangledown\mathcal{H}(\mathbf{Z}^{(m,k)}) \\
    &= \left[(1-3\alpha)\mathbf{I} + \alpha\mathbf{S}^{(m,k)}\right]\mathbf{Z}^{(m,k)}
    +  \alpha\mathbf{P}^{(k)}_{em}\mathbf{Z}^{(e,k)}+ \alpha\mathbf{Z}^{(m)}_{init},
\end{aligned}
\end{equation}

\begin{theorem}\label{theorem1}
 Let $\mathbf{Z}^{(m,k)}$ follow the update rule in Equation \eqref{first-order}. A sufficient condition for ensuring monotonic non-increasing behavior of the objective function $\mathcal{H}\left(\{\mathbf{Z}^{(m,k)}\}_{m=1}^M\right)$:

\[\mathcal{H}(\{\mathbf{Z}^{(m,k+1)}\}_{m=1}^M) \leq \mathcal{H}(\{\mathbf{Z}^{(m,k)}\}_{m=1}^M),\]
is that the step size $\alpha$ satisfies:

\[0 < \alpha \leq \min_m \left\{2 \cdot \|3\mathbf{I}-\mathbf{S}^{(m,k)}-\mathbf{P}^{(k)}_{mm}\mathbf{1}^{N\times N}\|_2^{-1}\right\}. \]
\end{theorem}
See \textbf{Appendix A} for proof of the theorem and other details.
Theorem \ref{theorem1} indicates that ensuring a consistent decrease in the objective function may require complementary constraints on the step sizes. In such cases, excessively small step sizes result in slow convergence, whereas overly large step sizes can lead to divergence and instability. 
Furthermore, adapting the step size to the $M$ omics datasets is both computationally demanding and cumbersome in terms of selection.

To address this issue, we employ the second-order Newton method, thereby circumventing the complexities involved in step size adjustment.
The details are as follows:

\begin{equation} \label{reverse}
 \begin{aligned}
    &\mathbf{Z}^{(m,k+1)} =\mathbf{Z}^{(m,k)}- [\triangledown^2 \mathcal{H}(\mathbf{Z}^{(m)})]^{-1}\triangledown\mathcal{H}(\mathbf{Z}^{(m,k)}) \\
    &=\mathbf{Z}^{(m,k)} - [ 3\mathbf{I} -(\mathbf{S} ^{(m,k)}+\mathbf{P}^{(k)}_{mm}\mathbf{1}^{N\times N})]^{-1} \triangledown\mathcal{H}(\mathbf{Z}^{(m,k)}),
\end{aligned}
\end{equation}
where $\triangledown^2 \mathcal{H}(\mathbf{Z}^{(m)})$ is the Hessian matrix. This formulation dynamically adjusts the step size based on the local curvature of the objective function.
Given that directly computing the matrix inverse for large-scale Hessian matrices is computationally prohibitive, we approximate the inverse using the Neumann series expansion. For $\rho(\frac{\mathbf{S} ^{(m,k)}}{3}) < 1$, the inverse can be approximated by truncating the series after the first term as:
\begin{equation} \label{approximation}
\begin{aligned}
\left[ 3\mathbf{I} -(\mathbf{S} ^{(m,k)}+\mathbf{P}^{(k)}_{mm}\mathbf{1}^{N\times N})\right]^{-1}    \approx
 \frac{1}{3}\left(\mathbf{I} +\frac{\mathbf{S} ^{(m,k)}}{3}\right)
\end{aligned}.
\end{equation}
Substituting the approximated inverse Hessian into Equation \eqref{reverse} results in:
\begin{equation} \label{second-order}
 \begin{aligned}
  &\mathbf{Z}^{(m,k+1)} 
  =\mathbf{Z}^{(m,k)}- \frac{1}{3}\left(\mathbf{I} +\frac{\mathbf{S} ^{(m,k)}}{3}\right)
  \triangledown\mathcal{H}(\mathbf{Z}^{(m,k)})\\
  & =\frac{\mathbf{S}^{(m,k)^2}}{9}\mathbf{Z}^{(m,k)}+ \frac{3\mathbf{I}+\mathbf{S}^{(m,k)}}{9}(\sum_{e=1}^M  \mathbf{P}^{(k)}_{em}\mathbf{Z}^{(e,k)}
  + \mathbf{Z}^{(m)}_{init}).
\end{aligned}
\end{equation}

For the equation above, we introduce alternative terms to simplify it. The final propagation steps are shown as follows:
\begin{equation}
 \begin{aligned}
\mathbf{S}_{ij}^{(m,k)}&=1+\left\langle\frac{\mathbf{W}_K^{(m,k)}\mathbf{Z}_i^{(m)}}{\|\mathbf{W}_K^{(m,k)}\mathbf{Z}_i^{(m)}\|_2},\frac{\mathbf{W}_Q^{(m,k)}\mathbf{Z}_j^{(m)}}{\|\mathbf{W}_Q^{(m,k)}\mathbf{Z}_j^{(m)}\|_2}\right\rangle,\\
    \mathbf{P}_{em}^{(k)} &=  1 + \exp \left(-
    \left({\mathbf{W}^{(k)}_{K}\mathbf{Z}^{(e)}}\right)^\top
    \otimes \left({\mathbf{W}^{(k)}_{Q}\mathbf{Z}^{(m)}}\right) \right),
\end{aligned}
\end{equation}
where $\mathbf{W}^{(m,k)}_{K}$, $\mathbf{W}^{(m,k)}_{Q}$, $\mathbf{W}^{(k)}_{K}$, and $\mathbf{W}^{(k)}_{Q}$ are learnable weight matrix in the $k$-layer, $\otimes$ denotes the tensor operation.  

\begin{theorem}\label{theorem2}
If the update rule in Equation \eqref{second-order} is followed, the objective function $\mathcal{H}\left(\{\mathbf{Z}^{(m,k)}\}_{m=1}^M\right)$ exhibits monotonic non-increasing behavior across iterations, i.e., 
\[\mathcal{H}\left(\{\mathbf{Z}^{(m,k+1)}\}_{m=1}^M\right) \leq \mathcal{H}\left(\{\mathbf{Z}^{(m,k)}\}_{m=1}^M\right), \quad \forall k \geq 0.\]
\end{theorem}
See \textbf{Appendix A} for proof of the theorem and other details.
To compute the doubly stochastic symmetric matrix $\mathbf{P}$, we employ a cyclic constraint projection method known as Dykstra’s algorithm, which is proven to converge for projections onto the nonempty intersection of finitely many closed convex sets.
The central idea is to decompose the entire constraint set into multiple simpler subsets, each allowing an easily determined projection operator.
\begin{equation} 
 \begin{aligned}
    &\mathcal{C}_1 = \{\mathbf{P}|\mathbf{P} = \mathbf{P}^\top\}, \quad \text{(Symmetry)} \\
    &\mathcal{C}_2 = \{\mathbf{P}|\mathbf{P}\mathbf{1} =\mathbf{1}\},\quad \text{(Row~Normalization)}\\
    &\mathcal{C}_3 = \{\mathbf{P}|\mathbf{P}^\top\mathbf{1} =\mathbf{1}\},\quad \text{(Column~Normalization)}
\end{aligned}
\end{equation}
where $\mathbf{1}$ is all one vector.
More details about the iteration process as shown in Appendix A. 
For each modality, we perform a $K$-step data update to obtain $\{\mathbf{Z}^{(m, K)}\}^{M}_{m=1}$, after which we integrate this modality into a fused representation to make the final decision.
\begin{equation} 
 \begin{aligned}
 \mathbf{Z} &= \mathrm{FUSE}(\{\mathbf{Z}^{(m,K)}\}^{M}_{m=1})\\
   \tilde{\mathbf{Y}} &= \mathrm{Softmax}(\mathbf{Z}\mathbf{W}),
\end{aligned},
\end{equation}
where $\mathbf{W}$ is the learnable weight and the $\mathrm{FUSE(\cdot)}$ denotes the fuse function, which usually can be average, sum, concatenate, and attention operation.
For a semi-supervised classification task, the proposed method employs a loss function defined by the cross-entropy loss:
\begin{equation}\label{crossentropy}
	\mathcal{L}_{Ce} =  -\sum_{i \in \Phi }\sum_{j=1}^c \mathbf{Y}_{ij}ln\tilde{\mathbf{Y}}_{ij},
\end{equation}
where $\Phi$ is the set of samples with labels. The total loss of the proposed method is:
\begin{equation}\label{total}
	\mathcal{L}_{\mathrm{total}} = \mathcal{L}_{Ct}+\mathcal{L}_{Ce}.
\end{equation}

The time complexity for aligning data from various modalities is $\mathcal{O}(MN^2d + MNd^2)$.
The time complexity for forward propagation is $\mathcal{O}(K(MNd^2 + M^2Nd))$.
Therefore, the total time complexity of the system is $\mathcal{O}(MN^2d + K(MNd^2 + M^2Nd))$.

\begin{table*}[!ht]
  \centering
  \setlength{\tabcolsep}{7.6pt} 
  \renewcommand{\arraystretch}{1.2} 
  \begin{tabular}{
    >{\centering\arraybackslash}m{1.0cm}| 
    >{\centering\arraybackslash}m{1.8cm}|| 
    >{\centering\arraybackslash}m{1.4cm} 
    >{\centering\arraybackslash}m{1.4cm} 
    >{\centering\arraybackslash}m{1.4cm} 
    >{\centering\arraybackslash}m{1.4cm} 
    >{\centering\arraybackslash}m{1.6cm} 
    >{\centering\arraybackslash}m{1.4cm}   
    >{\centering\arraybackslash}m{1.4cm}
  }
    \toprule
    {\textbf{Metrics}} & {\textbf{Methods/ Datasets}} & \textbf{BRCA} & \textbf{KIPAN} & \textbf{LGG} & \textbf{UCEC} & \textbf{CDRD} & \textbf{GBMLGG}& \textbf{TCGA} \\
    \midrule
    \multirow{8}{*}{\textbf{ACC}} 
      & \textbf{SVM}       & 70.2 (0.0) & 93.9 (0.0) & 60.0 (0.0) & 72.5 (0.0) & 76.3 (0.0) &  52.0 (0.0)  &  68.0 (0.0)\\
      & \textbf{RF}        & 80.2 (0.0) & 93.9 (0.0) & 49.8 (0.0) & 72.5 (0.0) & 77.6 (0.0) & \cellcolor{blue!10}54.6 (0.0)  & 67.0 (0.0)\\
      & \textbf{DeepMO}    & \cellcolor{blue!10}81.4 (2.1) & \cellcolor{blue!10}94.1 (0.9) & \cellcolor{blue!10}64.2 (3.1) & 80.2 (3.1) & 75.4 (4.7) & 53.5 (2.4)  &73.0 (6.9)\\
      & \textbf{MOGONET}   & 71.6 (1.5) & 91.4 (0.5) & 62.3 (1.8) & 75.4 (1.9) & 75.9 (0.0) & 49.2 (1.5)  &42.1 (0.7)\\
      & \textbf{MoGCN}      & 73.6 (0.4) & 91.5 (0.4) & 51.1 (0.0) & 71.6 (0.0) & \cellcolor{blue!10}77.9 (1.2) & 53.0 (6.3) & \cellcolor{blue!10}73.9 (0.4)\\
      & \textbf{Moanna}     & 77.5 (1.8) & 92.5 (0.9) & 61.3 (1.8) & 81.6 (2.2) & 62.1 (2.9) & 53.1 (4.9) & 58.7 (2.2)\\
      & \textbf{MOSGAT}     & 73.0 (3.2) & 91.1 (0.4) & 58.5 (1.3) & \cellcolor{blue!10}82.1 (1.9) & 75.7 (1.1) & 49.6 (1.2)&- \\
      & \textbf{GTMancer}       & \cellcolor{red!10}84.4 (1.4) & \cellcolor{red!10}94.8 (0.1) & \cellcolor{red!10}68.7 (0.6) & \cellcolor{red!10}84.2 (1.4) & \cellcolor{red!10}82.9 (1.2) & \cellcolor{red!10}56.5 (0.7) & \cellcolor{red!10}74.1 (0.5)\\
    \midrule
    \multirow{8}{*}{\textbf{F1}} 
      & \textbf{SVM}       &  43.6 (0.0) & \cellcolor{blue!10}92.9 (0.0) & 60.0 (0.0) & 28.0 (0.0) & 43.3 (0.0) &  39.4 (0.0) & 61.4 (0.0)\\
      & \textbf{RF}        & 68.8 (0.0) & 92.8 (0.0) & 49.6 (0.0) & 28.7 (0.0) & 48.9 (0.0) & 47.0 (0.0) & 52.0 (0.0)\\
      & \textbf{DeepMO}    & \cellcolor{blue!10}76.4 (4.9) & 92.5 (0.8) & \cellcolor{blue!10}63.7 (3.8) & \cellcolor{blue!10}53.8 (1.9)  & 67.6 (5.2) & \cellcolor{blue!10}51.1 (2.2) & 64.6 (6.9)\\
      & \textbf{MOGONET}   & 58.9 (2.6) & 90.5 (0.5) & 61.8 (2.4) & 43.7 (0.6)           & 43.2 (0.0) & 43.6 (2.7) &38.5 (0.4)\\
      & \textbf{MoGCN}      & 55.3 (0.7) & 91.4 (0.4) & 33.8 (0.0) & 28.0 (0.0)           &\cellcolor{blue!10}67.7 (4.6) & 41.5 (8.4) & \cellcolor{blue!10}66.7 (0.4)\\
      & \textbf{Moanna}     & 69.9 (2.0) & 90.3 (1.7) & 60.8 (1.9) & 53.4 (2.0)    & 58.5 (2.5) & 48.1 (3.0) & 51.7 (2.3)\\
      & \textbf{MOSGAT}     & 57.8 (5.3) & 90.8 (0.5) & 55.0 (2.9) & 52.3 (4.1)           & 47.8 (3.1) & 46.7 (1.5) &-\\
      & \textbf{GTMancer}       & \cellcolor{red!10}80.0 (1.6) & \cellcolor{red!10}93.9 (0.8) & \cellcolor{red!10}68.9 (0.6) & \cellcolor{red!10}57.6 (1.8) & \cellcolor{red!10}69.4 (1.9) & \cellcolor{red!10}53.4 (0.7) &  \cellcolor{red!10}66.9 (0.3)\\
    \bottomrule
  \end{tabular}  
  \caption{Classification results (mean\% and standard deviation\%) of all comparative algorithms on the seven cancer subtype datasets, supervised by 10\% of the labeled samples, where the best results are filled with \sethlcolor{red!10}\hl{\textbf{red}} and the second-best results are filled with \sethlcolor{blue!10}\hl{\textbf{blue}}.}
    \label{performance}
\end{table*}

\section{Experiment}
We evaluate the performance of our model by applying it to different types of cancer datasets.
Specifically, our study aims to address the following questions:
\begin{itemize}
    \item (RQ1) How does GTMancer perform relative to state-of-the-art methods for handling multi-omics datasets in cancer research?
    \item (RQ2) Can GTMancer maintain its effectiveness in categorizing cancer subtypes when faced with extreme scarcity of labeled samples?
    \item (RQ3) Are the alignment and dynamic information exchange mechanisms integral to improving cancer subtype identification accuracy?
    \item (RQ4) How sensitive is the proposed method to variations in the temperature parameter $\tau$?
\end{itemize}
\subsection{Datasets}
The study utilizes seven multi-omics datasets across various cancer types, including BRCA, KIPAN, LGG, UCEC, CDRD, GBMLGG, and TCGA. Each dataset incorporates three or two types of omics data, such as DNA methylation, mRNA, and miRNA (or other data types like CNV and RPPA in specific cases), and is used for classifying cancer subtypes or grades. 
The more details of these datasets are shown in \textbf{Appendix B.1}.

\subsection{Compared Methods}
We evaluate the proposed method against seven benchmark algorithms, comprising both traditional machine learning techniques and multi-omics analysis approaches. The machine learning algorithms include Support Vector Machines (SVM) and Random Forests (RF), while the multi-omics methods consist of DeepMO \cite{DeepMo}, MOGONET \cite{wang2021mogonet}, MoGCN \cite{Li2022MOGCN}, Moanna \cite{lupat2023moanna}, and MOSGAT \cite{wu2024mosgat}.
All compared methods use the default parameters following the original paper.  
The more details are shown in \textbf{Appendix B.2}.

\subsection{Experimental Setting}
For the proposed method, we specify the following hyperparameters: The layers number $K=3$, the temperature $\tau = 10$, the learning rate is set as $1e-2$, the training epoch is $200$, the weight decay set as $5e-5$, and the random dropout is $0.5$.
In multi-omics semi-supervised classification, the proposed method leverages a split of 10\% supervised omics samples for training and 90\% unsupervised samples for testing.  
Due to the need for additional labeled samples to create a validation set, we do not use a separate validation set. 
Instead, we evaluate the model using the parameters from the final iteration for testing.
More experiment results are shown in \textbf{Appendix C}.

\begin{figure}[!ht] 
	\centering
	\includegraphics[width=0.48\textwidth]{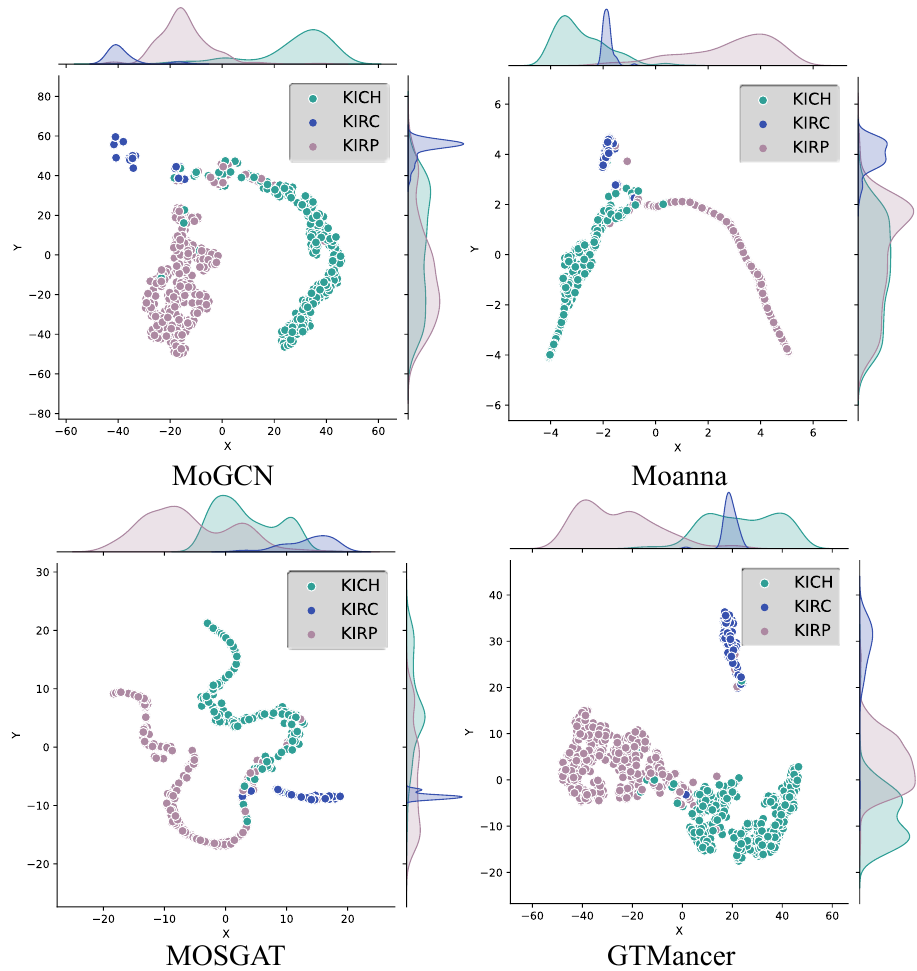}\\
	\caption{The visualization of the representations of three compared methods on the KIPAN dataset.}
        \label{TSNE}
\end{figure}
\begin{figure}[!ht] 
	\centering
	\includegraphics[width=0.47\textwidth]{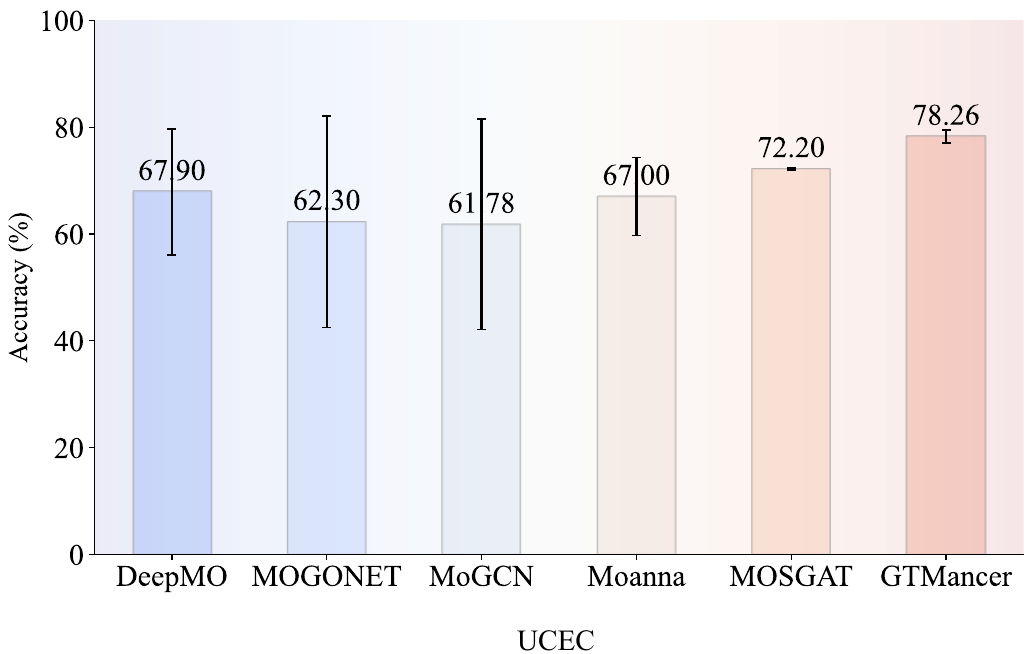}\\
	\caption{Performance comparison of the proposed GTMancer with other algorithms on the UCEC dataset under extremely few labeled samples (1\% label ratio).}
    \label{Ratio}
\end{figure}
\subsection{Cancer Subtype Classification (RQ1)}
As shown in Table \ref{performance}, our method outperforms the seven comparative algorithms in both ACC and F1 scores across most datasets, particularly the BRCA and KIPAN datasets. 
In contrast, traditional machine learning methods such as SVM and RF exhibit inferior performance on multiple datasets. 
While multi-omics deep learning approaches like DeepMO and MOGONET achieve competitive results on certain datasets, they still lag behind our method overall. 
Our approach demonstrates high robustness and adaptability, maintaining excellent performance across datasets of varying complexity.
It significantly enhances ACC and F1 metrics, underscoring its superior advantage in multi-omics data analysis.

\begin{figure*}[!ht] 
	\centering
	\includegraphics[width=0.97\textwidth]{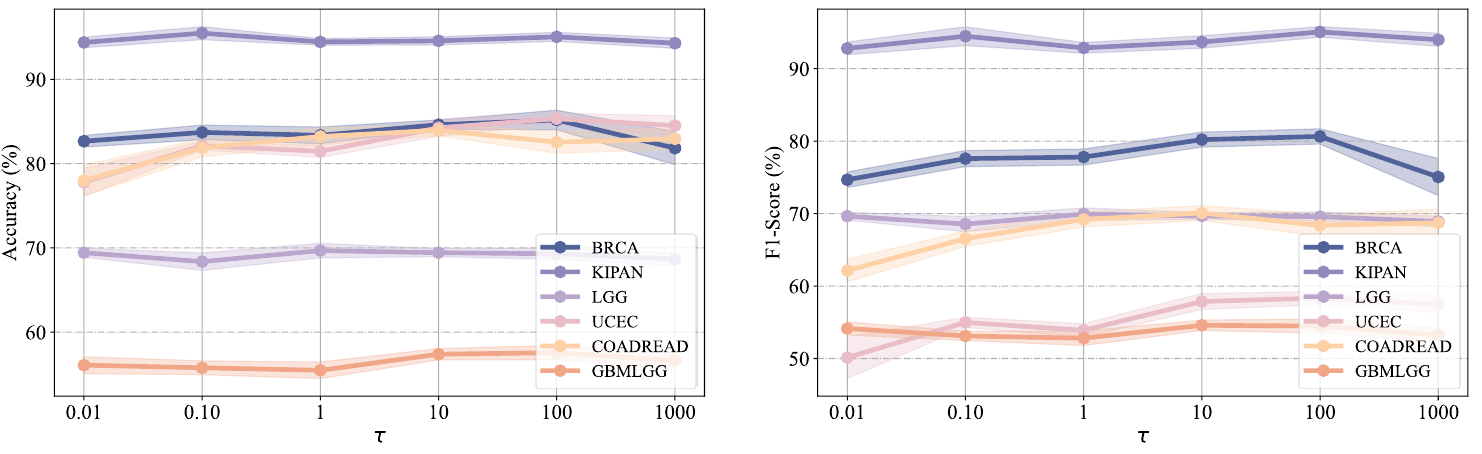}\\
	\caption{Performance of the proposed method under different temperature coefficients ($\tau$).}
    \label{Sensitivity}
\end{figure*}
\begin{figure}[!ht] 
	\centering
	\includegraphics[width=0.48\textwidth]{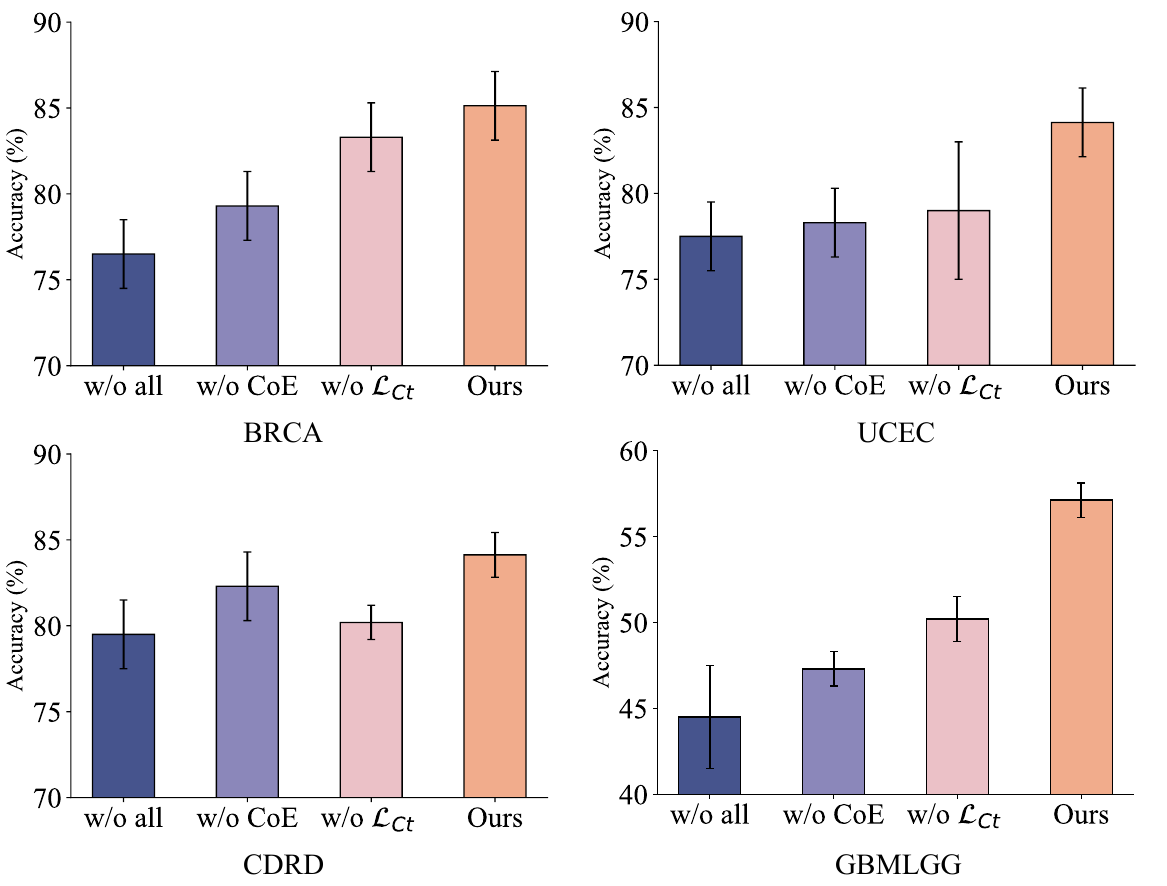}\\
	\caption{Ablation study of the proposed GTMancer framework on four datasets.}
    \label{Ablantion}
\end{figure}

\subsection{Visualization (RQ1)}
To further validate the effectiveness of the proposed method in multi-omics tasks, we conducted t-SNE visualizations on multi-omics representations derived from various approaches.
As shown in Figure \ref{TSNE},
the embeddings derived from GTMancer exhibit highly distinct clusters for the three categories (labels KICH, KIRC, and KIRP), with minimal overlap between the clusters. 
This distinct separation is particularly evident in the marginal distributions along both the X and Y axes, where the distributions of different categories are compact and well-separated.
In contrast, the other methods (MoGCN, Moanna, and MOSGAT) show varying degrees of cluster overlap, particularly for categories KIRC and KIRP, which are less distinguishable.
This performance underscores the robust feature extraction and precise category representation capabilities of our method.

\subsection{Performance in few labeled scenarios (RQ2)}
In this subsection, we further analyze the performance of GTMancer in minimally labeled scenarios. 
As shown in Figure \ref{Ratio}, the proposed GTMancer is highly effective in scenarios of extreme sample scarcity. Furthermore, the proposed method has smaller fluctuations.
By harnessing the powerful integration and alignment of diverse omics information, it significantly outperforms state-of-the-art methods. This makes the framework particularly well-suited for applications with limited labeled data, such as rare diseases or resource-constrained studies.

\subsection{Ablation Study (RQ3)}
To further validate the effectiveness of the proposed modules, we developed three experimental variants: w/o ALL, w/o $\mathcal{L}_{Ct}$, and w/o CoE (Cross-omics Exchange).
As shown in Figure \ref{Ablantion}, the ablation study results demonstrate that the complete model consistently outperforms all variants across all datasets, particularly on complex datasets such as BRCA and GBMLGG.
This underscores the crucial role of each module's synergy in enhancing model performance. 
Removing all modules (w/o ALL) led to a significant decline in performance, while eliminating specific modules (w/o $\mathcal{L}_{Ct}$ or w/o CoE) also resulted in reduced performance, especially on complex datasets.
In contrast, on simpler datasets like KIPAN and CDRD, the impact of module removal was minimal, indicating that the feature relationships in these datasets are more straightforward.
Overall, the complete model exhibits high adaptability.

\subsection{Parameter Sensitivity (RQ4)}
In this section, we perform a sensitivity analysis on the temperature coefficient $\tau$ in the model, with the results presented in Figure \ref{Sensitivity}.
The value of $\tau$ directly influences the feature alignment between different modalities. 
A smaller $\tau$ (e.g., 0.01 or 0.1) produces an overly smooth similarity distribution, weakening the contrast between positive and negative samples.
This impairs modality alignment and results in lower overall ACC and F1 scores.
Conversely, a larger $\tau$ (e.g., 1000) makes the similarity distribution overly sharp, causing the model to overemphasize high-similarity positive samples while neglecting global information from negative samples, ultimately degrading performance.
A moderate $\tau$ (e.g., 10, 100) achieves a balance between positive and negative sample contrast, effectively capturing global features across modalities and emphasizing key connections, which leads to optimal performance for most cancer types.

\section{Conclusion}
In this paper, we proposed the GTMancer framework, which provided multi-omics data with a global receptive field across modalities, resulting in improved performance. 
Inspired by graph optimization processes, we designed a multi-omics graph optimization objective and derived the forward propagation formula using gradient descent. 
This approach captures both omics-omics and sample-sample associations through an attention mechanism operating within and among omics.
The convergence of this optimization process was rigorously proven. 
To further mitigate the effects of step size during gradient descent, we employed Newton's method as an alternative and also provided theoretical proof of its convergence.
Experimental results demonstrated that our framework outperforms existing state-of-the-art methods, highlighting its effectiveness and competitiveness.
\section*{Acknowledgements}
This work was supported by the National Natural Science Foundation of China (Grant Nos. 62202422 and 62372408).

\section*{Contribution Statement}
Jielong Lu and Zhihao Wu contributed equally as co-first authors.

\bibliographystyle{named}
\bibliography{ijcai25}

\end{document}